\documentclass[11pt]{article}
 
\usepackage{arxiv}

\usepackage{hyperref}       
\usepackage{url}            
\usepackage{booktabs}       
\usepackage{amsfonts}       
\usepackage{nicefrac}       
\usepackage{microtype}      
\usepackage{graphicx}
\usepackage[numbers]{natbib}
\usepackage{doi}

\usepackage{multirow}
\usepackage{algorithm}
\usepackage{algpseudocode}
\usepackage{graphicx}
\usepackage{xcolor}
\usepackage{soul}
\usepackage{amsmath}
\usepackage{amsfonts}
\usepackage{caption,subcaption}
\usepackage[para]{footmisc}
\usepackage{booktabs}
\usepackage{xcolor,colortbl}
\usepackage{hyperref}
\hypersetup{colorlinks = true, citecolor = blue }
\usepackage{cleveref}
\usepackage{caption}
\newcommand\blfootnote[1]{%
  \begingroup
  \renewcommand\thefootnote{}\footnote{#1}%
  \addtocounter{footnote}{-1}%
  \endgroup
}

\usepackage[utf8]{inputenc}
\usepackage[LFE,LAE,T1]{fontenc}
\usepackage{arabtex}
\usepackage{utf8}

\title{Arabic Sentiment Analysis with Noisy Deep Explainable Model}

\author{ Md Atabuzzaman$^{\dagger,1}$, Md Shajalal$^{\dagger,2,3}$,Maksuda Bilkis Baby$^{4}$. Alexander Boden$^{2,5}$\\
    $^{1}$Virginia Tech, Virginia, USA \\
    $^{2}$Fraunhofer Institute for Applied Information Technology FIT, Germany\\
    $^{3}$University of Siegen, Germany\\
    $^{4}$UITS, Dhaka, Bangladesh\\
    $^{5}$Bonn-Rhein-Sieg University of Applied Sciences, Germany
}

\begin{document}
\maketitle

\textbf{Abstract.}
Sentiment Analysis (SA) is an essential task for numerous real-world applications. However, the majority of SA research focuses on high-resource languages such as English and Chinese, while limited-resource languages like Arabic and Bengali receive less attention. Additionally, existing Arabic sentiment analysis methods based on advanced artificial intelligence (AI) approaches tend to operate as black boxes, making it challenging to comprehend the reasoning behind their predictions. This paper proposes an explainable sentiment classification framework for the Arabic language. We introduce a noise layer to different deep learning (DL) models, including BiLSTM and CNN-BiLSTM, to address the issue of overfitting. The proposed framework enables the explanation of specific predictions by training a local surrogate explainable model, shedding light on the reasons behind each sentiment prediction (positive or negative). Experiments were conducted on publicly available benchmark Arabic SA datasets, and the results demonstrated that the inclusion of noise layers in DL model improves performance for the Arabic language by mitigating overfitting. Our method also outperformed several state-of-the-art approaches. Moreover, the introduction of explainability with the noise layer enhances transparency and accountability, making the model suitable for practical adoption in AI-enabled systems.\blfootnote{$\dagger$Both authors contributed equally \\This is the pre-print version of our accepted paper at the 7th International Conference on Natural Language Processing and Information Retrieval~(ACM NLPIR'2023)}


\section{Introduction}
Online social media platforms have become increasingly popular, leading to the emergence of various fields dedicated to analyzing the platforms and their content in order to extract useful information for individuals~\cite{villalobos2022sentimental}. Sentiment analysis~(SA) is one of them. It is a branch of Natural Language Processing~(NLP) which is concerned with identifying the feelings expressed in texts. However, SA begins to thrive merely due to the emergence of the Web, which makes it possible to include interactive content. This implies that people are free to upload any kind of content, including their own ideas and beliefs. SA may indeed be used to investigate this enormous volume of raw text data in order to provide a concise summary of what the public believes about a specific topic or a product, or even about any opinion~\cite{liu2012sentiment,al2019comprehensive}.

Prior research on sentiment analysis mostly focused on high-resourced languages. Since Arabic is not a high-resourced language, still there is a lack of attention as compared to other high-resourced languages. The prior methods for Arabic Sentiment Analysis~(ASA) depended on sentiment lexicons like ArSenL~\cite{badaro2014large}, a large-scale MSA word lexicon. Various options for analyzing Arabic-specific data were examined using recurrent and recursive neural networks~\cite{al2015deep,al2017aroma,baly2019arsentd}. Natural language procesing tasks got new dimension in terms of accuracy after the introduction of word-embedding techniques. The textual representation with pre-trained word-embedding trained on multiple large corpus were used for sentiment analysis tasks. Dahou et al.~\cite{dahou2019arabic} trained CNN with the semantic representation using word-embedding for analysing sentiment on Arabic text. Farha et al.~\cite{farha2019mazajak} proposed a hybrid model for ASA by employing LSTMs for sequence and context interpretation and CNNs for feature extraction. Then a BERT-based model for Arabic language representation, AraBERT was proposed for many Arabic language-specific tasks including ASA~\cite{antoun2020arabert}. 

However, the proposed models work like a black-box, even the developers and AI practitioners do not exactly understand what are the causes for a specific prediction~(positive or negative). This lack of transparency is a drawback for an efficient ASA system for adoption in real-world applications. Though some works can be found with explanations for SA prediction using XAI tool-kits for rich languages such as English~\cite{ribeiro2016should,mathews2019explainable}, to the best of our knowledge, there is no single work of Arabic sentiment where XAI is utilized to explain the reasons of prediction of the complex models. In addition, deep learning~(DL)-based models often show over-fitting characteristics due to the lack of sufficient amount of data used to train~\cite{altowayan2016word} including the Arabic SA task. This reduces the models' efficiency to determine the sentiment of low-resources languages like Arabic, Bengali, Hindi, etc.  

To tackle these concerns, we propose a new interpretable\footnote{The words \emph{Explainability} and \emph{Interpretability} are used interchangeably thought the paper.} Arabic sentiment classification framework by adding a Gaussian noise layer to the DL-based models. We develop and train two DL models including Bidirectional LSTM (BiLSTM) and CNN-BiLSTM, CNN layer followed by BiLSTM layer for sentiment classification. The experiments results indicate that adding noise layer helps to resolve the overfitting problem of these models in SA. To explain particular prediction of our sentiment classification framework, we adopted LIME~(Local Interpretable Model-agnostic Explanations), a prominent XAI method that can explain the predictions of any sentiment classifier in an interpretable and transparent manner, by learning an interpretable surrogate model locally around the prediction~\cite{ribeiro2016should}. 
For experimental purpose, we employed publicly available datasets including a Large Arabic Book Review~(LABR)~\cite{aly2013labr} and an Arabic Hotel Review~(HTL) datasets~\cite{elsahar2015building}. The experimental results verify our claims and mitigate above-mentioned concerns reducing potential overfitting problem for DL-based ASA models. The contribution of this paper can be summarized as follows:
\begin{enumerate}
    \item{We propose two different DL-based methods by introducing \emph{noise layer} for Arabic SA to reduce over-fitting with improved performance.}
    \item{To the best of our knowledge, this is the initial endeavor towards enhancing the explainability of Arabic sentiment classification models.}
    \item{Our method consistently achieves competitive performance compared to state-of-the-art approaches in Arabic SA, and it can be applied to other regional Arabic languages.}
\end{enumerate}
    
The rest of the paper is organized as follows: We survey related literature on ASA in Section~\ref{related works}. Then we present our method with explainability in Section~\ref{proposed techniques}. In Section~\ref{result and discussion}, we discuss the findings from the experiments. Finally, Section \ref{conclusion} concludes our methods with some future plans.

\section{Literature Review} \label{related works}
Recently, like other languages, Arabic SA had gained the attention of the research community~\cite{villalobos2022sentimental}. Farra et al.~\cite{farra2010sentence} worked with SA utilizing Arabic sentence structure with grammatical approach and lexicon-based approach. Then Abdul-Mageed et al.~\cite{abdul2011subjectivity} proposed methods to identify subjectivity and sentiment of standard Arabic. In the following years they proposed a corpus for sentiment analysis and a system to detect social media post's sentiment. In these works, the authors utilized a large set of features for the experiments with machine learning algorithms~\cite{farha2021comparative}. Shoukry et al.~\cite{shoukry2012sentence} classified the sentiment of Egyptian Arabic tweets using SVM and NB classifiers. Later, they measured the performance of different machine learning models on preprocessed (i.e., stemming, stop words removal and normalization) tweets~\cite{shoukry2015hybrid}. Duwairi et al.~\cite{duwairi2014sentiment} also employed classical ML-models including NB, K-Nearest Neighbors (KNN) and SVM classifiers to perform SA on Jordanian Arabic tweets. Nayel et al.\cite{nayel2021machine} employed a classical machine learning algorithm including SVM~(Support Vector Machine) for sentiment and sarcasm detection purposes.

However, the use of DL methods is less common in Arabic SA compared to English. An LSTM-CNN model is utilized by Sarah \cite{alhumoud2015hybrid} for Arabic  text to classify two unbalanced classes from the ASTD dataset among four classes. Similarly, a CNN model is used by \cite{oueslati2020review} with Stanford segmenter for the purposes of tweets tokenization and normalization. They applied the CNN model on the ASTD dataset with a word-embedding model. Heikal et al.~\cite{heikal2018sentiment} proposed a method combining CNN and LSTM models with pre-trained word-embedding model to predict the sentiment of the tweets. Some more prominent works on ASA also came out from workshop and shared tasks~\cite{habash2021proceedings}. Hengle et al.~\cite{hengle2021combining} combined context-free and contextualized representations for Arabic sarcasm detection and sentiment analysis. Word-embedding model was also applied in multiple works~\cite{zahran2015word, atabuzzaman2021leveraging,shajalalsentence,shajalal2016query,shajalal2020coverage} including sentiment analysis, textual similarity estimation and intent mining.

As BERT~(Bidirectional Encoder Representations from Transformers) based models show very promising performance in English SA, Oueslati et al.~\cite{eljundi2019hulmona} presented an Arabic language-specific universal language model~(ULM), hULMonA by fine-tuning multi-lingual BERT~(mBERT). To evaluate the ULM, they collected a benchmark dataset for SA. Safaya et al.~\cite{safaya2020bert} proposed an ArabicBERT model by utilizing a pre-trained BERT model~(bert-base-arabic) with CNNs. Another BERT-based Arabic language representation model, AraBERT is developed by \cite{antoun2020arabert} to improve the state-of-the-art in many Arabic natural language understanding tasks. Wadhawan et al.~\cite{wadhawan2021arabert} tried to classify the sentiment using the segmentation based method and two transformer based methods. Husain et al.~\cite{husain2021leveraging} hypothesized that tweets are more likely to contain offensive content when the tweet is positive or negative. Therefore, they fine-tuned the AraBERT using offensive language for Arabic sarcasm detection and sentiment analysis.

However, XAI is being applied for SA analysis for some languages like English and Chinese~\cite{ribeiro2016should,danilevsky2020survey}. Though many works have been done on ASA, they did not employ XAI for explaining the reasons for a specific prediction of the used ML or DL models. Adding a noise layer to the DL-based models also helps reducing over-fitting and eventually enhance the performance of the DL models~\cite{you2019adversarial,neelakantan2015adding}. In addition, there is no such work on ASA where a noise layer is added in the DL models to reduce over-fitting and improve the model's performance with XAI.

\begin{figure}[!h]
    \centering
    \includegraphics[width=0.8\linewidth]{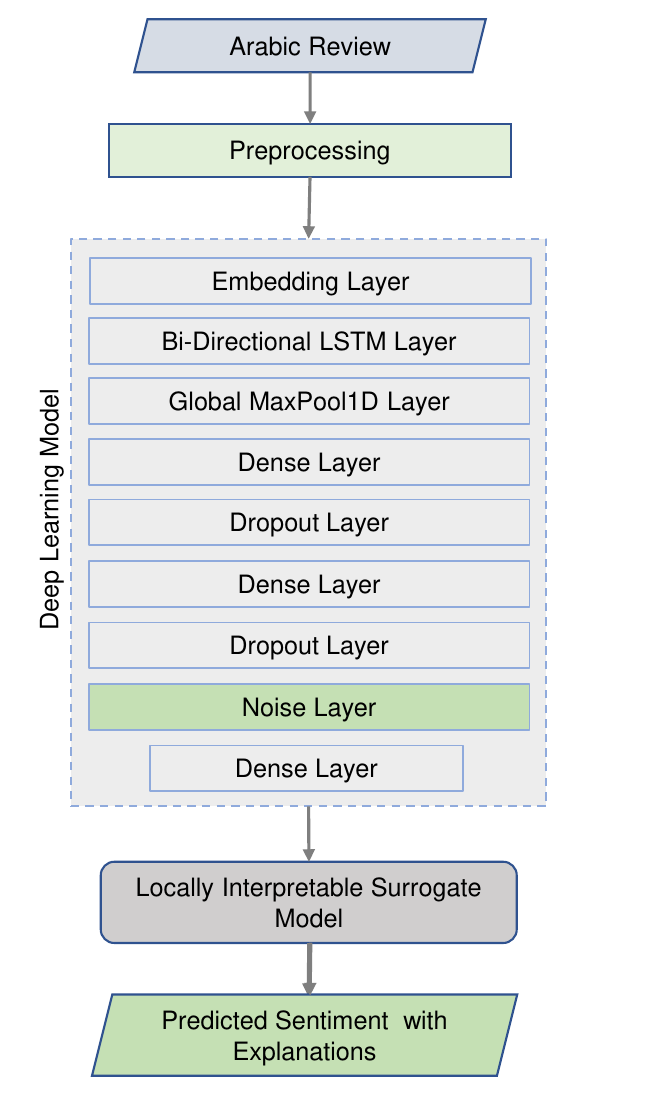}
    \caption{Overview diagram of our proposed Arabic Sentiment Analysis method}
    \label{overview}
\end{figure} 

\section{Proposed Method} \label{proposed techniques}
This section presents our proposed method for Arabic Sentiment classification, which utilizes DL and provides explanations for each prediction that attempt to highlight the reasons for certain prediction. Initially, the Arabic reviews serve as input for the DL methods. Subsequently, the reviews undergo a preprocessing phase, where special characters are removed. Following this, a word tokenizer is employed to extract the list of words from the reviews. Arabic words are then assigned indexes, resulting in a sequential representation that mirrors the order of words in the reviews. In situations where some reviews have a smaller number of words, a padding technique is applied to ensure uniformity in the sequences. Consequently, the padded sequences of the reviews are obtained from the preprocessing phase. BiLSTM and CNN-BiLSTM models are trained using the padded sequences of the reviews. The performance of these trained models is evaluated using test data. Finally, to explain specific predictions, a locally trained surrogate model is employed, utilizing LIME. An overview of the overall framework is depicted in Fig.~\ref{overview}.

\subsection{Classification Models} 
In the BiLSTM model, an Embedding layer is followed by a bidirectional LSTM layer and a global max pooling layer. In the CNN-BiLSTM model, an additional convolutional layer is introduced before the BiLSTM layer. Subsequently, two dense layers with dropout layers are appended after the max pooling layer, with each dense layer for both model. To mitigate overfitting, a noise layer is incorporated just before the final output layer in both case. The inclusion of the noise layer is motivated by the need to address overfitting in small neural networks trained with limited training data~\cite{you2019adversarial}. Inspired by this approach, we integrate a Gaussian noise layer into both the BiLSTM and CNN-BiLSTM models to mitigate overfitting~\cite{you2019adversarial,neelakantan2015adding}. The detail for each layer is mentioned in section~\ref{ES}.
\subsection{Explainable Surrogate Model with LIME}
For the explainability purpose, we trained a local surrogate model that mimics the classification performance of the original model. In this regard, we apply LIME~(Local Interpretable Model-agnostic Explanations) introduced by~\cite{ribeiro2016should}. LIME is a local delegate model means it is a trained model used to explain the causes of the predictions of the underlying black-box complex structure. However, it also includes generating different versions of the data for the machine learning model and testing what happens to the predictions, utilizing this perturbed data as a training set instead of the initial training data. In another sense, LIME creates a new dataset using permuted data and the associated black-box model predictions. LIME then learns an explainable model on this new dataset, which is weighted by the sampled instances' closeness to the instance of interest.

The computation of the surrogate model can be defined as follows:  
\begin{equation}
    \xi(x) = \underset{g \in \mathbf{G}}{\mathrm{arg \: min}} \:\: \textit{L}(f,g,\pi_{x'}) + \Omega(g),
\end{equation}
where $g$ represents an explanation model for an instance $x$, $G$ represents an explanation family, $f$ is the original model (i.e., our BiLSTM model), and $\textit{L}$ is the loss function. Model complexity is $\Omega(g)$. LIME explains local predictions of the model.

\section{Experimental Results with Discussion} \label{result and discussion}
\subsection{Dataset}
For experiment and verification purposes, we apply our methods on two Arabic benchmark datasets for sentiment analysis including LABR~\cite{aly2013labr} and Hotel review Dataset (HIL)~\cite{elsahar2015building}.
\\

\noindent \textbf{LABR Dataset:} Large-scale Arabic Book Review (LABR) dataset contains \emph{63257} reviews each with a rating of 1~(one) to 5~(five) on \emph{2131} books by \emph{16486} users. Reviews with rating 4 and 5 are considered as positive~(1) sentiment and rating with 1 and 2 are counted as negative~(0) sentiment~\cite{aly2013labr}. We eliminate the reviews with rating 3 as they indicate neutral reviews~\cite{aly2013labr}. Then \emph{51056} reviews are considered for our experiments among them 42832 are positive~(1) and \emph{8224} are negative~(0). The unequal no. of 1 and 0 makes it a imbalance dataset. Finally, \emph{40844} reviews are used for training purposes and the rest \emph{10212} are for testing. 
\\

\noindent \textbf{Hotel Review dataset~(HTL):} This dataset contains reviews written in the Arabic language. The reviews are for 8100 Hotels by 13K users~\cite{elsahar2015building}. A total of \emph{15572} reviews are there in the HTL dataset. Among~(HTL)~\cite{elsahar2015building} that are written in the Arabic language. The reviews are for \emph{8100} Hotels among them \emph{10766} reviews are positive, \emph{2645} are Negative and the rest are Neutral. No. of positive reviews is almost four times higher than no. negative reviews, which makes it an imbalanced dataset. To make it a balanced and smaller dataset, we only select \emph{2645} positive reviews randomly. Finally, \emph{3967} reviews are used for training and the rest \emph{1323} reviews for testing.

\begin{table*}[ht]
    \centering
    \small
    \caption{Performance of BiLSTM and CNN-BiLSTM models on imbalance LABR dataset}
    \label{LABR result}
    \begin{tabular}{|c|c|c|c|c|c|c|c|c|} \hline
      \multirow{2}{*}{model} &\multirow{2}{*}{setup} &\multirow{2}{*}{$train_-acc$} &\multicolumn{4}{|c|}{Testing Evaluation}  &\multirow{2}{*}{$test_-acc$}&\multirow{2}{*}{overfit~(\%)}\\ \cline{4-7}
       &&&class &precision    &recall  &f1-score &&\\ \hline \hline

          \multirow{6}{*}{BiLSTM} &\multirow{2}{*}{$Model_{ND}$} & \multirow{2}{*}{0.98} &0       &0.62      &0.55      &0.58  &\multirow{2}{*}{0.88} &\multirow{2}{*}{10.0}\\ 
         & & &1       &0.92      &0.94      &0.93 & & \\ \cline{2-9}

          &\multirow{2}{*}{$Model_{N}$} &\multirow{2}{*}{0.99} &0       &0.63      &0.53      &0.57  &\multirow{2}{*}{0.88} &\multirow{2}{*}{11.0}\\ 
         & & &1       &0.92      &0.94      &0.93  &&\\ \cline{2-9}
 
         &\multirow{2}{*}{$Model_{D}$} &\multirow{2}{*}{0.98} &0       &0.56      &0.59      &0.57  &\multirow{2}{*}{0.86} &\multirow{2}{*}{12.0}\\ 
         & & &1       &0.92      &0.91      &0.92  & &\\ \hline \hline

        \multirow{6}{1.5cm}{CNN-BiLSTM} &\multirow{2}{*}{$Model_{ND}$} &\multirow{2}{*}{0.99} &0       &0.63      &0.50      &0.56  &\multirow{2}{*}{0.88} &\multirow{2}{*}{11.0}\\ 
         & & &1       &0.91      &0.95      &0.93 & & \\ \cline{2-9}

         &\multirow{2}{*}{$Model_{N}$} &\multirow{2}{*}{0.99} &0       &0.62      &0.50      &0.55  &\multirow{2}{*}{0.88} &\multirow{2}{*}{11.0}\\ 
         & & &1       &0.91      &0.94      &0.93  & &\\ \cline{2-9}

         &\multirow{2}{*}{$Model_{D}$} &\multirow{2}{*}{0.99} &0       &0.60      &0.55      &0.57  &\multirow{2}{*}{0.87} &\multirow{2}{*}{12.0}\\ 
         & & &1       &0.92      &0.93      &0.93  & &\\ \hline
    \end{tabular}
    
\end{table*}

\begin{table}[!ht]
    \centering
    \caption{Performance of BiLSTM and CNN-BiLSTM models with different settings on balance HTL dataset}
    \begin{tabular}{|c|c|c|c|c|} \hline
    Model &Setup &Tr.-Acc.(\%) &Tst-Acc.(\%) &O.fit.~(\%) \\ \hline \hline
    \multirow{3}{*}{BiLSTM} &$Model_{ND}$ &\textbf{99.72} &\textbf{93.64} &\textbf{6.08}  \\ \cline{2-5}
     &$Model_N$ &99.86 &93.51 &6.35 \\ \cline{2-5}
     &$Model_D$ &99.96 &92.00 &7.96 \\ \hline \hline
    \multirow{3}{*}{CNN-BiLSTM} &$Model_{ND}$ &\textbf{99.85} &\textbf{94.78} &\textbf{5.07}  \\ \cline{2-5}
     &$Model_N$ &99.92 &93.65 &6.27 \\ \cline{2-5}
     &$Model_D$ &99.97 &92.02 &7.95 \\ \hline
    \end{tabular}  
    \label{result_table}
\end{table}

\subsection{Experimental setup} \label{ES}

We conducted experiments using different settings to evaluate the performance of our models. For each proposed model, we considered three variations, each with specific characteristics:

\begin{itemize}
\item $Model_{ND}$: This model includes both the noise and dropout layers before the output layer or final layer.
\item $Model_N$: This model only contains the noise layer before the output layer and does not have an immediate dropout layer before the noise layer.
\item $Model_D$: This model does not include the noise layer but incorporates the dropout layer.
\end{itemize}

For the BiLSTM model, the architecture follows the overview depicted in Figure~\ref{overview}. In this model, we enumerated 10,000 unique vocabularies, representing each vocabulary with a 100-dimensional vector using the embedding layer of the Keras library. Subsequently, four fully connected layers (dense layers) were implemented, with each layer containing 128, 64, 32, and 1 neuron, respectively. All dense layers utilized the ReLU activation function, except for the last layer, which employed the Sigmoid activation function.

For the CNN-BiLSTM model, a one-dimensional convolutional layer with a ReLU activation function and a kernel size of 3 was included. A dropout layer with a value of 0.5 was used to randomly drop 50\% of the features during training. Additionally, a Gaussian noise layer with a value of 0.75 was employed. The models were compiled using the Adam optimizer with the binary cross-entropy loss function, and accuracy was used as the evaluation metric. For training and testing, 80\% of the reviews were used for training, while the remaining 20\% were used for testing, with a batch size of 64. Each variant of the DL-based models underwent training for a total of 10 epochs. Finally, a local interpretable surrogate model was trained to mimic the original proposed models and provide explanations for specific predictions.

\subsection{Experimental results}

Tables~\ref{LABR result} and~\ref{result_table} present the performance of our BiLSTM and CNN-BiLSTM models in determining Arabic sentiment, using different evaluation metrics on the LABR and HTL datasets, respectively. As LABR is an imbalanced dataset, in addition to accuracy, other evaluation metrics are used to assess the performance of our proposed models. Conversely, as HTL is a balanced dataset, the performance of the introduced BiLSTM and CNN-BiLSTM models is evaluated solely based on accuracy.

We have two DL-based models, each with three different setups. When the noise layer is added along with the dropout layer before the last layer of the models, they exhibit improved performance and reduced overfitting. Table~\ref{LABR result} supports this claim by showing that when the noise layer is absent before the output layer~($Model_{D}$), the overfitting~(O.fit) is 12\%. However, after adding the noise layer before the output layer~($Model_{ND}$), the overfitting decreases by 2\% and 1\% for the BiLSTM and CNN-BiLSTM models, respectively. The \textit{precision} of negative reviews in the BiLSTM model's $Model_{ND}$ setup in Table~\ref{LABR result} is 0.62, while for $Model_D$, it is 0.56. This indicates that the noise layer helps in identifying more negative reviews compared to without the noise layer. The same trend is observed for the CNN-BiLSTM model

Table~\ref{result_table} illustrates that the overfitting~(O.fit.) of the BiLSTM model is 6.08\% when the noise and dropout layers are included~($Model_{ND}$). Similarly, when the noise layer and dropout layer are integrated into the CNN-BiLSTM model, the overfitting is 5.07\%, with training and testing accuracies of 99.85\% and 94.78\%, respectively. On the other hand, without the noise layer, the degree of overfitting increases. For example, in the CNN-BiLSTM model with the $Model_D$ setup, the overfitting is 7.95\%, which is approximately 2.88\% higher than $Model_{ND}$(5.07\%). These findings demonstrate the effectiveness of adding a noise layer to the models. When the dropout layer is removed before the output layer and the noise layer is added ($Model_N$), the overfitting between training accuracy and testing accuracy remains moderate, at 6.35\% for BiLSTM and 6.27\% for CNN-BiLSTM. These results further emphasize that adding a noise layer, either with or without the dropout layer, improves the performance of the models.

\begin{table}[!h]
    \centering
    \caption{Performance comparison of our method~CNN-BiLSTM with some known methods on LABR dataset.}
    \label{comparison_tablel}
    \begin{tabular}{lc} \hline
    methods &accuracy(\%)\\ \hline
    Attention-BiGRU~\cite{berrimi2023attention} &95.6\\
    SRU-Attention~\cite{al2019sentiment} &95.1\\
    AraBERT~\cite{antoun2020arabert} &89.6\\
    \textbf{CNN-BiLSTM}  &\textbf{88.0} \\ 
    Multi-chan CNN~\cite{dahou2019multi} &87.5\\
    mBERT~\cite{antoun2020arabert}   & 83.0\\
    CNN~\cite{elfaik2021deep} &81.6\\
    BiLSTM~\cite{elfaik2021deep} &80.7 \\
    LSTM~\cite{elfaik2021deep} &78.5\\
    SVM~\cite{elsahar2015building} &78.3\\ \hline
    \end{tabular}
\end{table}

\begin{figure*}[!h]
    \centering
    \includegraphics[width=.97\linewidth]{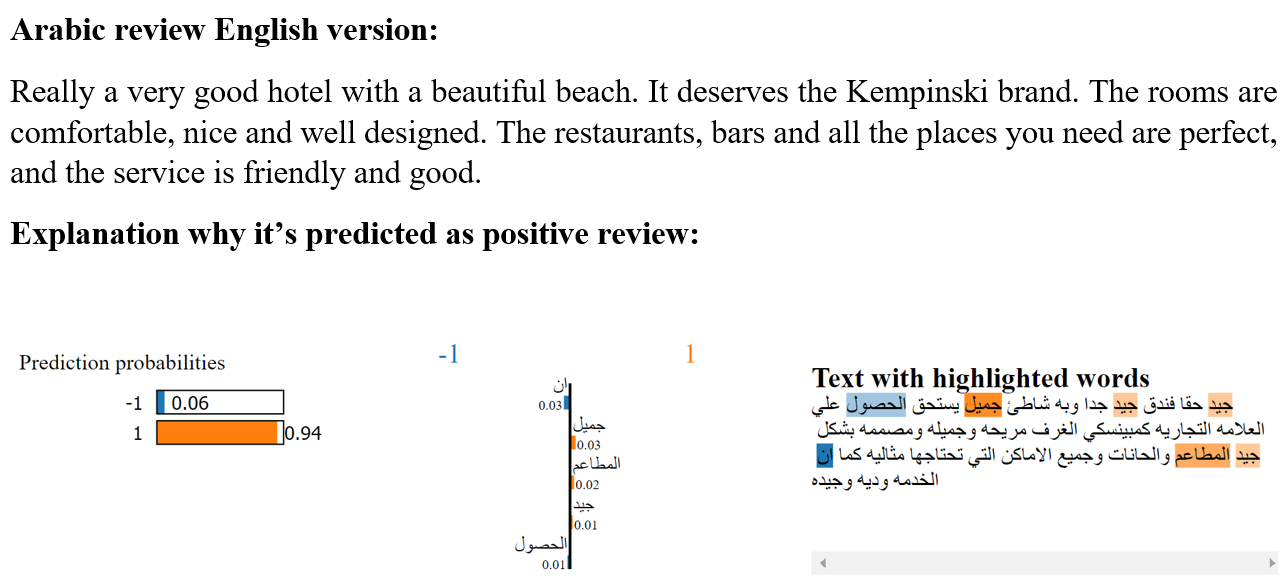}
    \caption{Explanation of a particular class~(Positive) using LIME.}
    \label{fig: positive}
\end{figure*}
\begin{figure*}[!htb]
    \centering
    \includegraphics[width=.97\linewidth]{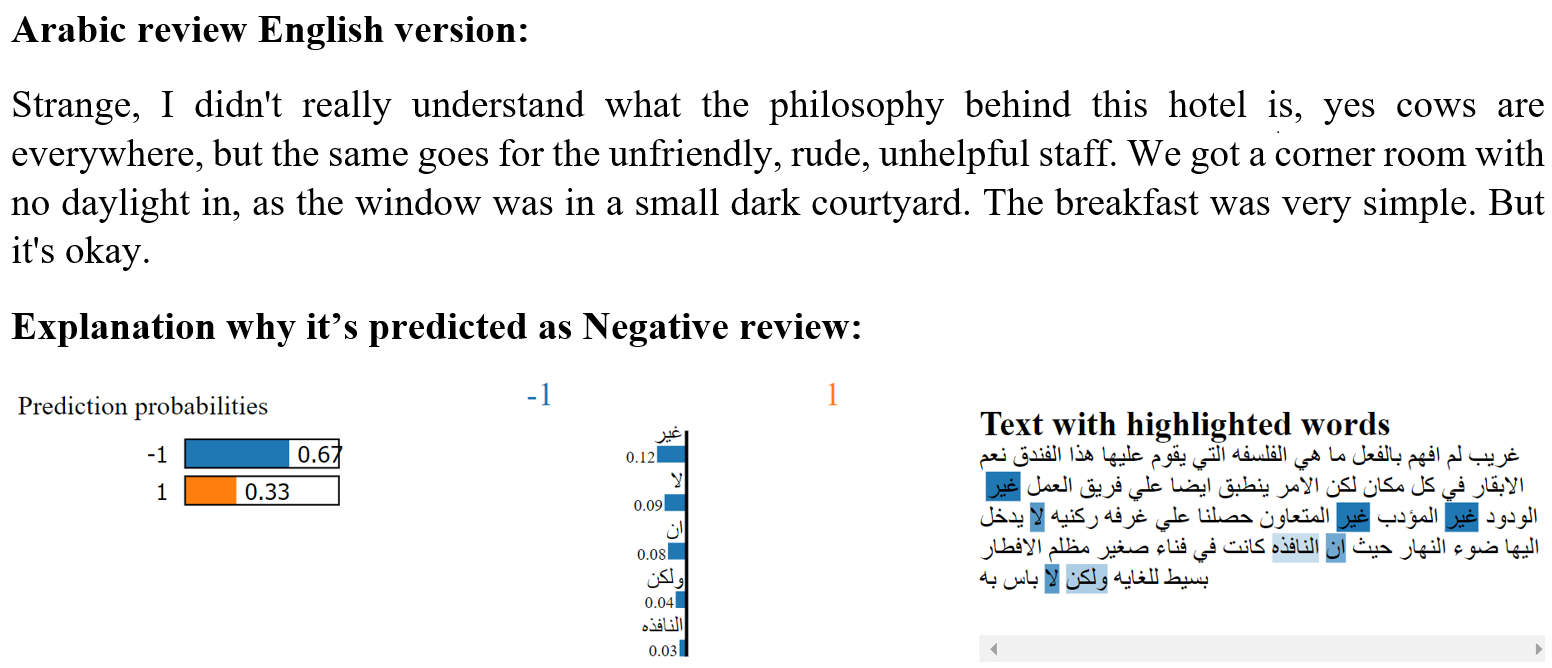}
    \caption{Explanation of a particular class~(Negative) using LIME.}
    \label{fig:negative}
\end{figure*}

\subsection{Comparison with related works}
Different prominent research works had been conducted on ASA on different dataset. Here, we compare proposed method with some existing methods that are on LABR dataset. Attention-BiGRU~\cite{berrimi2023attention}, SRU-Attention~\cite{al2019sentiment}, and AraBERT~\cite{antoun2020arabert} achieved higher accuracy than our proposed method as these are attention-based DL models. Attention BiGRU~\cite{berrimi2023attention} employed a hybrid bidirectional gated recurrent unit (BiGRU) and bidirectional
long short-term memory (BiLSTM) additive attention model with two types of embedding and achieved SOTA results on LABR dataset. SRU-Attention~\cite{al2019sentiment} used simple recurrent unit with attention mechanism and obtained an accuracy score of 95.1. AraBERT~\cite{antoun2020arabert} was introduced for different tasks including ASA. It is built especially for Arabic NLP. In ASA, they achieved 89.6\% accuracy which is 1.6\% more than our CNN-BiLSTM. Multilingual BERT~(mBERT) also employed for ASA~\cite{antoun2020arabert} and its accuracy is 83\% which is 5\% lesser than our method. Moreover, LSTM, BiLSTM and CNN, multi-channel CNN are also utilized without noise layer and their performance are quite promising but still the performance is lower than us. These findings illustrate the significant of adding noise layer in the DL models.  Though our proposed CNN-BiLSTM could not outperform the attention based methods, However it performed consistently for two dataset.

\begin{figure*}[!htb]
    \centering
    \includegraphics[width=.97\linewidth]{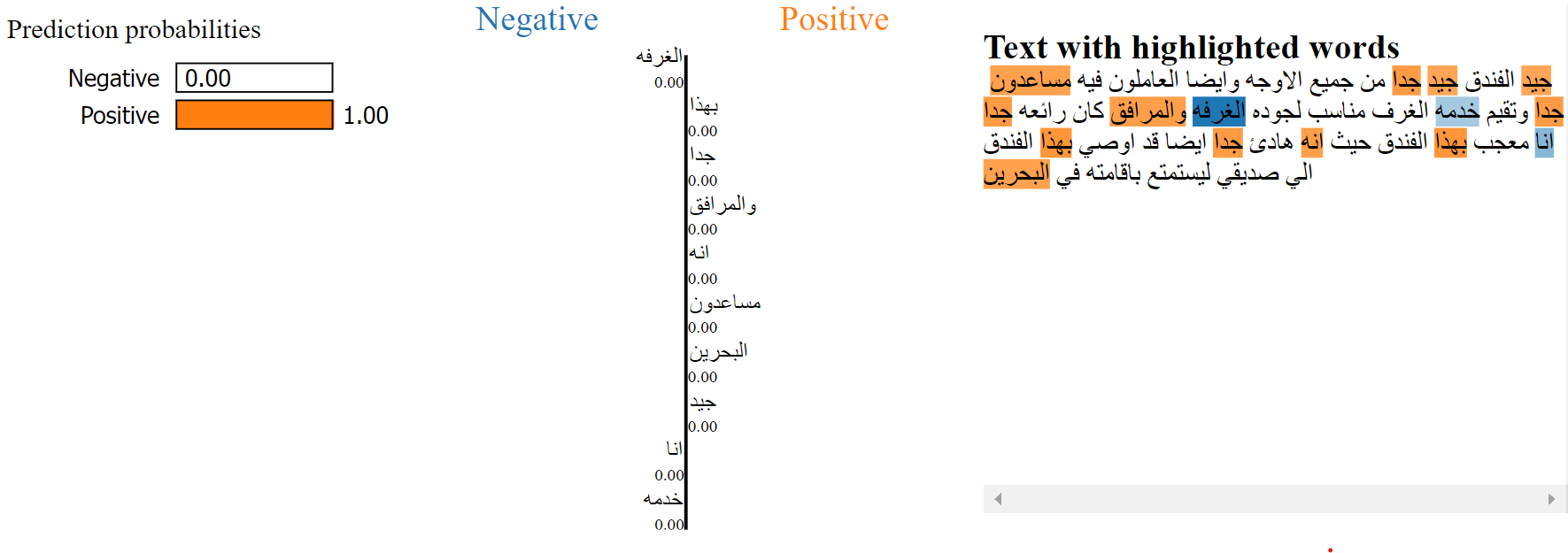}
    \caption{Explanation of a review with $Model_{ND}$ using LIME.}
    \label{fig: nd}
\end{figure*}
\begin{figure*}[!htb]
    \centering    \includegraphics[width=.97\linewidth]{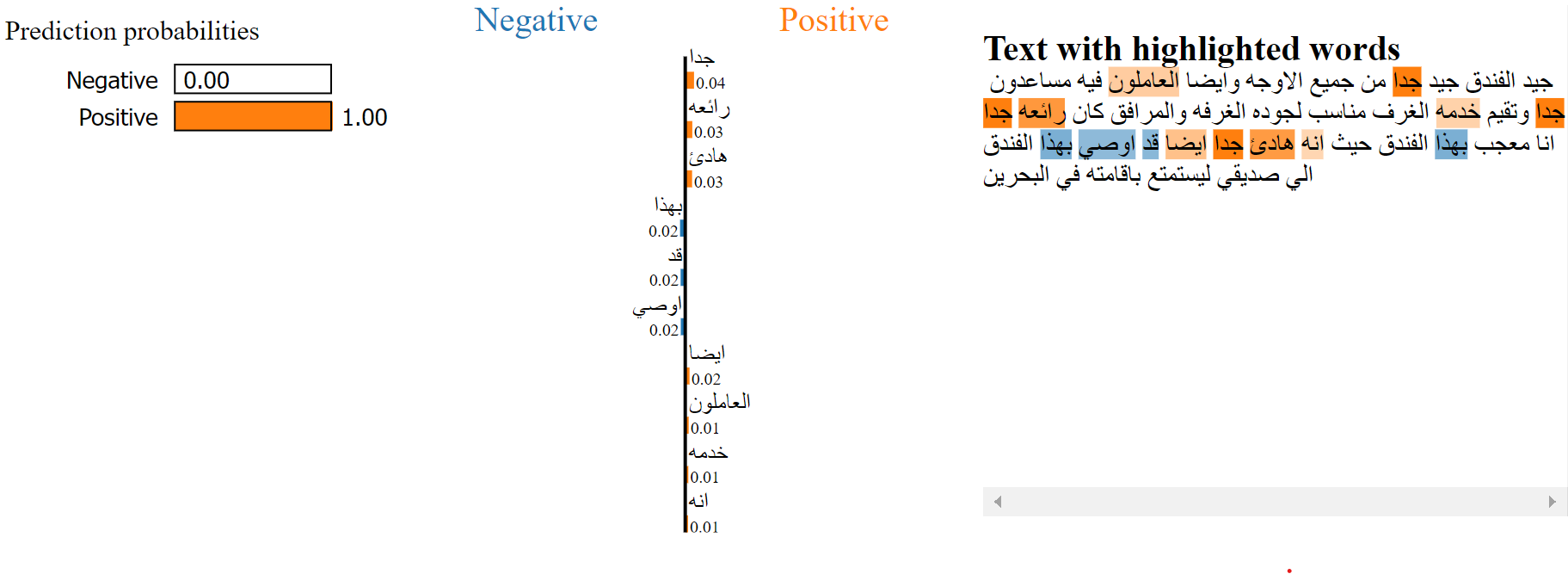}
    \caption{Explanation of a review with $Model_D$ using LIME.}
    \label{fig:d}
\end{figure*}

\subsection{XAI in Arabic Sentiment Analysis}
We trained a local explainable surrogate model to get an understandable representation of the predicted sentiments of the Arabic reviews. 

\if false
Fig.~\ref{fig: positive} and \ref{fig:negative} illustrate the humans interpretable representation of the predicted positive and negative sentiment, respectively. These figures present and highlight words that are contributing to the particular predictions. The other words which make the sentiment of the review positive~(1) are highlighted with orange color where deep orange means the words have more impact to predict the sentiment of the Arabic review as positive and light orange means the words have less impact to predict the review as positive. On the other hand, the words that are moving the review toward negative~(-1) are highlighted with blue color. In Fig.~\ref{fig: positive} the \texturdu{جميل}~(beautiful), \texturdu{جيد}~(good), etc. are the most significant words for predicting the review as positive and highlighted with orange color. In Fig.~\ref{fig:negative} the words \texturdu{ولكن}~(but), and \texturdu{غير}~(not), etc. are making the given review as negative sentiment and marked with blue color. The figures also show the predicted probabilities of the positive sentiment and negative sentiment. Each word's impact on a particular sentiment of the reviews is also illustrated in the middle of the figures. Anyone reading and seeing the explained representation of the reviews will be able to understand why the review is either positive or negative. It also makes our black-box mechanism of ASA understandable.
\fi

Let's consider an review and see how our proposed method perform on this. 
The review is as follow: ``\RL{جيد الفندق  جيد جدا من جميع الاوجه وايضا العاملون فيه مساعدون جدا وتقيم خدمه الغرفمناسب لجوده الغرفه والمرافق كان رائعه جدا انا معجب بهذا الفندق حيث انه هادئ جدا ايضا قد  اوصي بهذا الفندق الي صديقي ليستمتع باقامته في البحري}''~(\emph{Translation}: \textit{Good, the hotel is very good in all respects, and also the staff are very helpful, and the room service is appropriate for the quality of the room and the facilities were very wonderful. I admire this hotel as it is very quiet also. I may recommend this hotel to my friend to enjoy his stay in Bahrain.})

\noindent
Fig.~\ref{fig: nd} and \ref{fig:d} illustrate the humans interpretable representation of the predicted positive sentiment of the same review using $Model_{ND}$ and $Model_D$, respectively. Fig.~\ref{fig: nd} which is obtained using $model_{ND}$ highlights the words (e.g. \RL{جيد}~(good), \RL{جدا}~( very), \RL{هادئ}~(calm/quiet), \RL{مساعدون}~(helpful)) which are really contributing the review toward positive sentiment. On the other hand, LIME and $model_D$ interpreted the same review in Fig~\ref{fig:d} and it shows some major differences such as \RL{جيد}~(good), \RL{مساعدون}~(helpful) are not highlighted as positive sentiment words. However, these words have real impact to make the review as positive sentiment. This phenomenon illustrate the effectiveness of adding noise layer in the DL model to make it more explainable and acceptable. 

\section{Conclusion and Future Work} \label{conclusion}
This paper proposed explainable Arabic sentiment classification framework introducing noise layer in deep learning models including BiLSTM, and CNN-BiLSTM. Generally, DL-based models show overfitting characteristics when a small amount of data is used for training which makes the DL model's generalization capability poor. That is why a Gaussian noise layer is added to the proposed models to reduce overfitting and enhance performance. The experimental results also indicate that the noise layer helps to reduce the overfitting issue of the DL-based models and improve the performance. Again, these models work in a black-box manner in predicting the sentiment which is not understandable to humans. Therefore, to interpret the reasons for the particular sentiment predictions, a locally explainable surrogate model known as LIME is employed for the first time in this paper for ASA. LIME shows easy-to-understand explanations that provides an understandable representation to make users' sense for a prediction.

In the future, we plan to enhance the performance of the current explainable AI algorithms for a better understanding of ASA and other Arabic NLP tasks. We also would like to employ federated learning in ASA. 


\bibliographystyle{unsrtnat}
\bibliography{references}  

\begin{thebibliography}{43}
\providecommand{\natexlab}[1]{#1}
\providecommand{\url}[1]{\texttt{#1}}
\expandafter\ifx\csname urlstyle\endcsname\relax
  \providecommand{\doi}[1]{doi: #1}\else
  \providecommand{\doi}{doi: \begingroup \urlstyle{rm}\Url}\fi

\bibitem[Villalobos et~al.(2022)Villalobos, Forero, De~Mello, Valencia, Orjuela, Tanscheit, and Cavalcanti]{villalobos2022sentimental}
Cristian~Muoz Villalobos, Leonardo~Mendoza Forero, Harold De~Mello, Cesar Valencia, Alvaro Orjuela, Ricardo Tanscheit, and Marco~Pacheco Cavalcanti.
\newblock Sentimental analysis on social media comments with recurring models and pretrained word embeddings in portuguese.
\newblock In \emph{Proceedings of the 2022 6th International Conference on Natural Language Processing and Information Retrieval}, pages 205--209, 2022.

\bibitem[Liu(2012)]{liu2012sentiment}
Bing Liu.
\newblock Sentiment analysis and opinion mining.
\newblock \emph{Synthesis lectures on human language technologies}, 5\penalty0 (1):\penalty0 1--167, 2012.

\bibitem[Al-Ayyoub et~al.(2019)Al-Ayyoub, Khamaiseh, Jararweh, and Al-Kabi]{al2019comprehensive}
Mahmoud Al-Ayyoub, Abed~Allah Khamaiseh, Yaser Jararweh, and Mohammed~N Al-Kabi.
\newblock A comprehensive survey of arabic sentiment analysis.
\newblock \emph{Information processing \& management}, 56\penalty0 (2):\penalty0 320--342, 2019.

\bibitem[Badaro et~al.(2014)Badaro, Baly, Hajj, Habash, and El-Hajj]{badaro2014large}
Gilbert Badaro, Ramy Baly, Hazem Hajj, Nizar Habash, and Wassim El-Hajj.
\newblock A large scale arabic sentiment lexicon for arabic opinion mining.
\newblock In \emph{Proceedings of the EMNLP 2014 workshop on arabic natural language processing (ANLP)}, pages 165--173, 2014.

\bibitem[Al~Sallab et~al.(2015)Al~Sallab, Hajj, Badaro, Baly, El-Hajj, and Shaban]{al2015deep}
Ahmad Al~Sallab, Hazem Hajj, Gilbert Badaro, Ramy Baly, Wassim El-Hajj, and Khaled Shaban.
\newblock Deep learning models for sentiment analysis in arabic.
\newblock In \emph{Proceedings of the second workshop on Arabic natural language processing}, pages 9--17, 2015.

\bibitem[Al-Sallab et~al.(2017)Al-Sallab, Baly, Hajj, Shaban, El-Hajj, and Badaro]{al2017aroma}
Ahmad Al-Sallab, Ramy Baly, Hazem Hajj, Khaled~Bashir Shaban, Wassim El-Hajj, and Gilbert Badaro.
\newblock Aroma: A recursive deep learning model for opinion mining in arabic as a low resource language.
\newblock \emph{ACM Transactions on Asian and Low-Resource Language Information Processing (TALLIP)}, 16\penalty0 (4):\penalty0 1--20, 2017.

\bibitem[Baly et~al.(2019)Baly, Khaddaj, Hajj, El-Hajj, and Shaban]{baly2019arsentd}
Ramy Baly, Alaa Khaddaj, Hazem Hajj, Wassim El-Hajj, and Khaled~Bashir Shaban.
\newblock Arsentd-lev: A multi-topic corpus for target-based sentiment analysis in arabic levantine tweets.
\newblock \emph{arXiv preprint arXiv:1906.01830}, 2019.

\bibitem[Dahou et~al.(2019{\natexlab{a}})Dahou, Elaziz, Zhou, and Xiong]{dahou2019arabic}
Abdelghani Dahou, Mohamed~Abd Elaziz, Junwei Zhou, and Shengwu Xiong.
\newblock Arabic sentiment classification using convolutional neural network and differential evolution algorithm.
\newblock \emph{Computational intelligence and neuroscience}, 2019, 2019{\natexlab{a}}.

\bibitem[Farha and Magdy(2019)]{farha2019mazajak}
Ibrahim~Abu Farha and Walid Magdy.
\newblock Mazajak: An online arabic sentiment analyser.
\newblock In \emph{Proceedings of the Fourth Arabic Natural Language Processing Workshop}, pages 192--198, 2019.

\bibitem[Antoun et~al.(2020)Antoun, Baly, and Hajj]{antoun2020arabert}
Wissam Antoun, Fady Baly, and Hazem Hajj.
\newblock Arabert: Transformer-based model for arabic language understanding.
\newblock \emph{arXiv preprint arXiv:2003.00104}, 2020.

\bibitem[Ribeiro et~al.(2016)Ribeiro, Singh, and Guestrin]{ribeiro2016should}
Marco~Tulio Ribeiro, Sameer Singh, and Carlos Guestrin.
\newblock " why should i trust you?" explaining the predictions of any classifier.
\newblock In \emph{Proceedings of the 22nd ACM SIGKDD international conference on knowledge discovery and data mining}, pages 1135--1144, 2016.

\bibitem[Mathews(2019)]{mathews2019explainable}
Sherin~Mary Mathews.
\newblock Explainable artificial intelligence applications in nlp, biomedical, and malware classification: a literature review.
\newblock In \emph{Intelligent computing-proceedings of the computing conference}, pages 1269--1292. Springer, 2019.

\bibitem[Altowayan and Tao(2016)]{altowayan2016word}
A~Aziz Altowayan and Lixin Tao.
\newblock Word embeddings for arabic sentiment analysis.
\newblock In \emph{2016 IEEE International Conference on Big Data (Big Data)}, pages 3820--3825. IEEE, 2016.

\bibitem[Aly and Atiya(2013)]{aly2013labr}
Mohamed Aly and Amir Atiya.
\newblock Labr: A large scale arabic book reviews dataset.
\newblock In \emph{Proceedings of the 51st Annual Meeting of the Association for Computational Linguistics (Volume 2: Short Papers)}, pages 494--498, 2013.

\bibitem[ElSahar and El-Beltagy(2015)]{elsahar2015building}
Hady ElSahar and Samhaa~R El-Beltagy.
\newblock Building large arabic multi-domain resources for sentiment analysis.
\newblock In \emph{International conference on intelligent text processing and computational linguistics}, pages 23--34. Springer, 2015.

\bibitem[Farra et~al.(2010)Farra, Challita, Abou~Assi, and Hajj]{farra2010sentence}
Noura Farra, Elie Challita, Rawad Abou~Assi, and Hazem Hajj.
\newblock Sentence-level and document-level sentiment mining for arabic texts.
\newblock In \emph{2010 IEEE international conference on data mining workshops}, pages 1114--1119. IEEE, 2010.

\bibitem[Abdul-Mageed et~al.(2011)Abdul-Mageed, Diab, and Korayem]{abdul2011subjectivity}
Muhammad Abdul-Mageed, Mona Diab, and Mohammed Korayem.
\newblock Subjectivity and sentiment analysis of modern standard arabic.
\newblock In \emph{Proceedings of the 49th Annual Meeting of the Association for Computational Linguistics: Human Language Technologies}, pages 587--591, 2011.

\bibitem[Farha and Magdy(2021)]{farha2021comparative}
Ibrahim~Abu Farha and Walid Magdy.
\newblock A comparative study of effective approaches for arabic sentiment analysis.
\newblock \emph{Information Processing \& Management}, 58\penalty0 (2):\penalty0 102438, 2021.

\bibitem[Shoukry and Rafea(2012)]{shoukry2012sentence}
Amira Shoukry and Ahmed Rafea.
\newblock Sentence-level arabic sentiment analysis.
\newblock In \emph{2012 international conference on collaboration technologies and systems (CTS)}, pages 546--550. IEEE, 2012.

\bibitem[Shoukry and Rafea(2015)]{shoukry2015hybrid}
Amira Shoukry and Ahmed Rafea.
\newblock A hybrid approach for sentiment classification of egyptian dialect tweets.
\newblock In \emph{2015 First International Conference on Arabic Computational Linguistics (ACLing)}, pages 78--85. IEEE, 2015.

\bibitem[Duwairi et~al.(2014)Duwairi, Marji, Sha'ban, and Rushaidat]{duwairi2014sentiment}
Rehab~M Duwairi, Raed Marji, Narmeen Sha'ban, and Sally Rushaidat.
\newblock Sentiment analysis in arabic tweets.
\newblock In \emph{2014 5th international conference on information and communication systems (ICICS)}, pages 1--6. IEEE, 2014.

\bibitem[Nayel et~al.(2021)Nayel, Amer, Allam, and Abdallah]{nayel2021machine}
Hamada Nayel, Eslam Amer, Aya Allam, and Hanya Abdallah.
\newblock Machine learning-based model for sentiment and sarcasm detection.
\newblock In \emph{Proceedings of the Sixth Arabic Natural Language Processing Workshop}, pages 386--389, 2021.

\bibitem[Alhumoud et~al.(2015)Alhumoud, Albuhairi, and Alohaideb]{alhumoud2015hybrid}
Sarah Alhumoud, Tarfa Albuhairi, and Wejdan Alohaideb.
\newblock Hybrid sentiment analyser for arabic tweets using r.
\newblock In \emph{2015 7th International Joint Conference on Knowledge Discovery, Knowledge Engineering and Knowledge Management (IC3K)}, volume~1, pages 417--424. IEEE, 2015.

\bibitem[Oueslati et~al.(2020)Oueslati, Cambria, HajHmida, and Ounelli]{oueslati2020review}
Oumaima Oueslati, Erik Cambria, Moez~Ben HajHmida, and Habib Ounelli.
\newblock A review of sentiment analysis research in arabic language.
\newblock \emph{Future Generation Computer Systems}, 112:\penalty0 408--430, 2020.

\bibitem[Heikal et~al.(2018)Heikal, Torki, and El-Makky]{heikal2018sentiment}
Maha Heikal, Marwan Torki, and Nagwa El-Makky.
\newblock Sentiment analysis of arabic tweets using deep learning.
\newblock \emph{Procedia Computer Science}, 142:\penalty0 114--122, 2018.

\bibitem[Habash et~al.(2021)Habash, Bouamor, Hajj, Magdy, Zaghouani, Bougares, Tomeh, Farha, and Touileb]{habash2021proceedings}
Nizar Habash, Houda Bouamor, Hazem Hajj, Walid Magdy, Wajdi Zaghouani, Fethi Bougares, Nadi Tomeh, Ibrahim~Abu Farha, and Samia Touileb.
\newblock Proceedings of the sixth arabic natural language processing workshop.
\newblock In \emph{Proceedings of the Sixth Arabic Natural Language Processing Workshop}, 2021.

\bibitem[Hengle et~al.(2021)Hengle, Kshirsagar, Desai, and Marathe]{hengle2021combining}
Amey Hengle, Atharva Kshirsagar, Shaily Desai, and Manisha Marathe.
\newblock Combining context-free and contextualized representations for arabic sarcasm detection and sentiment identification.
\newblock \emph{arXiv preprint arXiv:2103.05683}, 2021.

\bibitem[Zahran et~al.(2015)Zahran, Magooda, Mahgoub, Raafat, Rashwan, and Atyia]{zahran2015word}
Mohamed~A Zahran, Ahmed Magooda, Ashraf~Y Mahgoub, Hazem Raafat, Mohsen Rashwan, and Amir Atyia.
\newblock Word representations in vector space and their applications for arabic.
\newblock In \emph{International Conference on Intelligent Text Processing and Computational Linguistics}, pages 430--443. Springer, 2015.

\bibitem[Atabuzzaman et~al.(2021)Atabuzzaman, Shajalal, Ahmed, Afjal, and Aono]{atabuzzaman2021leveraging}
Md~Atabuzzaman, Md~Shajalal, M~Elius Ahmed, Masud~Ibn Afjal, and Masaki Aono.
\newblock Leveraging grammatical roles for measuring semantic similarity between texts.
\newblock \emph{IEEE Access}, 9:\penalty0 62972--62983, 2021.

\bibitem[Shajalal and Aono()]{shajalalsentence}
Md~Shajalal and Masaki Aono.
\newblock Sentence-level semantic textual similarity using word-level semantics.

\bibitem[Shajalal et~al.(2016)Shajalal, Ullah, Chy, and Aono]{shajalal2016query}
Md~Shajalal, Md~Zia Ullah, Abu~Nowshed Chy, and Masaki Aono.
\newblock Query subtopic diversification based on cluster ranking and semantic features.
\newblock In \emph{2016 International Conference On Advanced Informatics: Concepts, Theory And Application (ICAICTA)}, pages 1--6. IEEE, 2016.

\bibitem[Shajalal and Aono(2020)]{shajalal2020coverage}
Md~Shajalal and Masaki Aono.
\newblock Coverage-based query subtopic diversification leveraging semantic relevance.
\newblock \emph{Knowledge and Information Systems}, 62:\penalty0 2873--2891, 2020.

\bibitem[ElJundi et~al.(2019)ElJundi, Antoun, El~Droubi, Hajj, El-Hajj, and Shaban]{eljundi2019hulmona}
Obeida ElJundi, Wissam Antoun, Nour El~Droubi, Hazem Hajj, Wassim El-Hajj, and Khaled Shaban.
\newblock hulmona: The universal language model in arabic.
\newblock In \emph{Proceedings of the fourth arabic natural language processing workshop}, pages 68--77, 2019.

\bibitem[Safaya et~al.(2020)Safaya, Abdullatif, and Yuret]{safaya2020bert}
Ali Safaya, Moutasem Abdullatif, and Deniz Yuret.
\newblock Bert-cnn for offensive speech identification in social media.
\newblock In \emph{Proceedings of the Fourteenth Workshop on Semantic Evaluation", Barcelona (online)", International Committee for Computational Linguistics}, 2020.

\bibitem[Wadhawan(2021)]{wadhawan2021arabert}
Anshul Wadhawan.
\newblock Arabert and farasa segmentation based approach for sarcasm and sentiment detection in arabic tweets.
\newblock \emph{arXiv preprint arXiv:2103.01679}, 2021.

\bibitem[Husain and Uzuner(2021)]{husain2021leveraging}
Fatemah Husain and Ozlem Uzuner.
\newblock Leveraging offensive language for sarcasm and sentiment detection in arabic.
\newblock In \emph{Proceedings of the Sixth Arabic Natural Language Processing Workshop}, pages 364--369, 2021.

\bibitem[Danilevsky et~al.(2020)Danilevsky, Qian, Aharonov, Katsis, Kawas, and Sen]{danilevsky2020survey}
Marina Danilevsky, Kun Qian, Ranit Aharonov, Yannis Katsis, Ban Kawas, and Prithviraj Sen.
\newblock A survey of the state of explainable ai for natural language processing.
\newblock \emph{arXiv preprint arXiv:2010.00711}, 2020.

\bibitem[You et~al.(2019)You, Ye, Li, Xu, and Wang]{you2019adversarial}
Zhonghui You, Jinmian Ye, Kunming Li, Zenglin Xu, and Ping Wang.
\newblock Adversarial noise layer: Regularize neural network by adding noise.
\newblock In \emph{2019 IEEE International Conference on Image Processing (ICIP)}, pages 909--913. IEEE, 2019.

\bibitem[Neelakantan et~al.(2015)Neelakantan, Vilnis, Le, Sutskever, Kaiser, Kurach, and Martens]{neelakantan2015adding}
Arvind Neelakantan, Luke Vilnis, Quoc~V Le, Ilya Sutskever, Lukasz Kaiser, Karol Kurach, and James Martens.
\newblock Adding gradient noise improves learning for very deep networks.
\newblock \emph{arXiv preprint arXiv:1511.06807}, 2015.

\bibitem[Berrimi et~al.(2023)Berrimi, Oussalah, Moussaoui, and Saidi]{berrimi2023attention}
Mohamed Berrimi, Mourad Oussalah, Abdelouahab Moussaoui, and Mohamed Saidi.
\newblock Attention mechanism architecture for arabic sentiment analysis.
\newblock \emph{ACM Transactions on Asian and Low-Resource Language Information Processing}, 22\penalty0 (4):\penalty0 1--26, 2023.

\bibitem[Al-Dabet and Tedmori(2019)]{al2019sentiment}
Saja Al-Dabet and Sara Tedmori.
\newblock Sentiment analysis for arabic language using attention-based simple recurrent unit.
\newblock In \emph{2019 2nd International Conference on new Trends in Computing Sciences (ICTCS)}, pages 1--6. IEEE, 2019.

\bibitem[Dahou et~al.(2019{\natexlab{b}})Dahou, Xiong, Zhou, and Elaziz]{dahou2019multi}
Abdelghani Dahou, Shengwu Xiong, Junwei Zhou, and Mohamed~Abd Elaziz.
\newblock Multi-channel embedding convolutional neural network model for arabic sentiment classification.
\newblock \emph{ACM Transactions on Asian and Low-Resource Language Information Processing (TALLIP)}, 18\penalty0 (4):\penalty0 1--23, 2019{\natexlab{b}}.

\bibitem[Elfaik et~al.(2021)]{elfaik2021deep}
Hanane Elfaik et~al.
\newblock Deep bidirectional lstm network learning-based sentiment analysis for arabic text.
\newblock \emph{Journal of Intelligent Systems}, 30\penalty0 (1):\penalty0 395--412, 2021.

\end{thebibliography}

\end{document}